\documentclass{article}

\usepackage[final,nonatbib]{neurips_2021}
\usepackage{url}            % simple URL typesetting
\usepackage{booktabs}       % professional-quality tables
\usepackage{amsfonts}       % blackboard math symbols
\usepackage{nicefrac}       % compact symbols for 1/2, etc.
\usepackage{microtype}      % microtypography
\usepackage{subcaption}
\usepackage{amsmath,amsfonts,amssymb}
\usepackage{breqn}
\usepackage{tabularx}
\usepackage{multirow}
\usepackage{graphicx}
\usepackage{color}
\usepackage{tcolorbox}
\usepackage{CJK}
\usepackage{adjustbox}
\usepackage{xcolor}
\usepackage{colortbl}
\usepackage{multicol}
\usepackage{vwcol} 
\newsavebox{\algleft}
\newsavebox{\algright}
\usepackage{bm}
\usepackage{booktabs}
\usepackage{ctable} % for \specialrule command
\usepackage{amsmath,stackengine}
\usepackage[linesnumbered,ruled,vlined]{algorithm2e}
\usepackage{times}
\usepackage{latexsym}
\usepackage{bbm}
\usepackage[utf8]{inputenc} % allow utf-8 input
\usepackage[T1]{fontenc}    % use 8-bit T1 fonts
\usepackage{hyperref}       % hyperlinks
\usepackage{url}            % simple URL typesetting
\usepackage{booktabs}       % professional-quality tables
\usepackage{amsfonts}       % blackboard math symbols
\usepackage{nicefrac}       % compact symbols for 1/2, etc.
\usepackage{microtype}      % microtypography
\usepackage{xcolor}         % colors

\title{{\color{black}Achieving Forgetting Prevention and Knowledge Transfer in Continual Learning}}

\author{
Zixuan Ke$^{1}$, Bing Liu$^{1}$, Nianzu Ma$^{1}$, Hu Xu$^{2}$ and Lei Shu$^{3}$\thanks{Work was done prior to joining Amazon.} \\ 
$^1$Department of Computer Science, University of Illinois at Chicago\\
$^2$Facebook AI Research\\
$^3$Amazon AWS AI\\
$^1$\texttt{\{zke4,liub,nma4\}@uic.edu}\\  $^2$\texttt{huxu@fb.com} \\ $^3$\texttt{shulindt@gmail.com}
}

\begin{document}

\maketitle
\begin{abstract}
{\color{black}Continual learning (CL) learns a sequence of tasks incrementally with the goal of achieving two main objectives: \textit{overcoming catastrophic forgetting} (CF) and \textit{encouraging knowledge transfer} (KT) \textit{across tasks}. However, most existing techniques focus only on overcoming CF and have no mechanism to encourage KT, and thus do not do well in KT. {\color{black}Although several papers have tried to deal with both CF and KT,} our experiments show that they suffer from serious CF when the tasks do not have much shared knowledge. {\color{black}Another observation is that most current CL methods do not use pre-trained models,} {\color{black}but it has been shown that such models can significantly improve the end task performance. For example, in natural language processing,  % \zixuan{the blue fonts text seems isolated with the next few sentences} 
fine-tuning a BERT-like pre-trained language model is one of the most effective approaches. However, for CL, this approach suffers from serious CF. An interesting question is how to make the best use of pre-trained models for CL. This paper proposes a novel model called CTR to solve these problems.} Our experimental results demonstrate the effectiveness of CTR.}\footnote{The code of CTR can be found at \url{https://github.com/ZixuanKe/PyContinual}} 
%(1) to overcome severe CF in BERT fine-tuning for continual learning, and (2) to transfer knowledge among tasks. CTR exploits BERT adapters to replace BERT fine-tuning to deal with BERT forgetting, and uses a capsule network with a knowledge transfer routing mechanism to enable cross task knowledge sharing. Experimental results demonstrate the effectiveness of CTR \footnote{The code has been uploaded as part of Supplementary.}

% {\color{red}Fine-tuning a BERT like language models using in-domain data is regarded as one of the most effective approaches for many natural language processing tasks. However, fine-tuning causes serious \textit{catastrophic forgetting} (CF) in \textit{continual learning}.~This is because fine-tuning exploits highly task specific features for good performance, but fine-tuning in learning a new task may completely destroy what has been tuned for previous tasks, resulting CF.
% This paper proposes a novel method called CTR (1) to overcome severe CF in BERT fine-tuning for continual learning, and (2) to transfer knowledge among tasks. CTR exploits BERT adapters to replace BERT fine-tuning to deal with BERT forgetting, and uses a capsule network with a knowledge transfer routing mechanism to enable cross task knowledge sharing. Experimental results demonstrate the effectiveness of CTR.}

\end{abstract}

%\zixuan{

%main paper: neurips_2021.tex; 

%Supplentary: neurips_2021_supplementary.tex;

% Previous author response: icml2021_author_response.tex}

\section{Introduction}
\label{sec.intro}

This paper studies continual learning (CL) of a sequence of natural language processing (NLP) tasks in the \textit{task continual learning} (Task-CL) setting. {\color{black}It aims to %\zixuan{is it two or three? do we want to include "how to better use pre-trained model"?} 
(i) prevent catastrophic forgetting (CF), and  % This is to avoid a negative case that tasks conflict with each other.
(ii) transfer knowledge across tasks.} (ii) is particularly important because many tasks in NLP share similar knowledge that can be leveraged to achieve better accuracy. CF means that in learning a new task, the existing network parameters learned for the previous tasks may be modified, which degrades the performance of previous tasks~\cite{mccloskey1989catastrophic}. In the Task-CL setting, the task id is provided for each test case in testing so that the specific model for the task in the network can be applied to classify the test case. Another popular CL setting is \textit{class continual learning}, which does not provide the task id during testing but it is for solving a different type of problems. 

Most existing CL papers focus on dealing with CF~\cite{ke2020mixed,chen2018lifelong}. There are also some papers that perform knowledge transfer. To achieve both objectives is highly challenging. To overcome CF in the Task-CL setting, we don't want the training of the new task to update the model parameters learned for previous tasks to achieve model separation. But to transfer knowledge across tasks, we want the new task to leverage the knowledge learned from previous tasks for learning a better model (forward transfer) and also want the new task to enhance the performance of similar previous tasks (backward transfer). This means it is necessary to update previous model parameters. This is a dilemma. {\color{black}Although several papers have tried to deal with both~\cite{ke2020continual,DBLP:conf/dasfaa/LvWLCZ19},} they were only tested using sentiment analysis tasks with strong shared knowledge. {\color{black}When tested with tasks that don't have much shared knowledge,} they suffer from severe CF (see Sec.~\ref{sec:results}). Those existing papers that focus on dealing with CF do not do well with knowledge transfer as they have no explicit mechanism to facilitate the transfer. % This paper attempts to achieve both objectives at the same time. 

% TIL trains one mixed model for all tasks in a sequence. In testing, the system is provided with the task ID for each test instance and the system uses only the model for the task to assign a class label to the test instance. In contrast, in CIL, the task ID is not provided but leave to the system to assign a class to the test instance.

%different topics of dialogues, different tasks of named-entity extractions
% , etc. Such shared knowledge should be exploited to produce better classifier in the CL setting than without CL.
% The focus of current CL research has been on dealing with CF (2)~\cite{Parisi2019continual}. 

{\color{black}Another observation about the current CL research is that most techniques do not use pre-trained models. But such pre-trained models or feature extractors can significantly improve the CL performance~\cite{hu2021continual,ke2021adapting}. An important question is how to make the best use of pre-trained models in CL. This paper studies the problem as well using NLP tasks, but we believe that the developed ideas are also applicable to computer vision tasks because most pre-trained models are based on the transformer architecture~\cite{vaswani2017attention}. We will see that the naive or the conventional way of directly adding the CL module on top of a pre-trained model is not the best choice (see Sec.~\ref{sec:results}).}

In NLP, fine-tuning a BERT~\cite{DBLP:conf/naacl/DevlinCLT19} like pre-trained language model has been regarded as one of the most effective techniques in applications~\cite{DBLP:conf/naacl/XuLSY19,sun-etal-2019-utilizing}. However, fine-tuning works poorly for continual learning. This is because the fine-tuned BERT for a task captures highly task-specific information~\cite{DBLP:journals/corr/abs-2004-14448}, which is difficult to be used by other tasks. When fine-tuning for a new task, it has to update the already fine-tuned parameters for previous tasks, which causes serious CF (see Sec.~\ref{sec:results}). %However \ls{What is interesting is that}, for the same sequence of tasks, if we replace BERT with word embeddings, there is little CF (see Sec.~\ref{sec:results}) because people use similar sentiment expressions in different domains/tasks, i.e., having significant knowledge sharing across tasks. However, using word embeddings gives lower accuracy for each task than using BERT fine-tuning. 
% If we want to use BERT without suffering from CF, we have to freeze BERT, which, however, gives lower accuracy. 

% \ls{what is our finding in Bert-finetune CL? what is the difference to other architecture's CL? We need these to explain why we use adapter.}

% does not produce the state-of-the-art accuracy results.

% that  Besides these two types of knowledge, it is also desirable to incorporate \textit{commonsense knowledge} that resides in pre-trained language models (such as BERT \cite{DBLP:conf/naacl/DevlinCLT19}) and how to adapt BERT for CL on downstream tasks is largely unknown to the community (our experiment shows that naively fine-tuning BERT for CL leads to poor performance.). 

% \bing{This section needs to be re-written to downplay Adapter and capsuleNet and emphasize transfer routing. Do we even want to call it Adapter as CL-plugin is very different from the original Adapter. Our purpose is also different. We do more than what the original Adapters does. In other words, we do more than just adaptation. In the adapter paper, my understanding is that each task has a adapter so that the tasks do not interfere with each other as they can all use the same underlying frozen BERT. In our case, we use only one ''adapter'' for all tasks. I suggest that we do not call CL-plugin an adapter. The term adapter only makes sense if for different tasks, we have different adapters. If all tasks use one ``adapter'', then it is no longer an adapter.} 

{\color{black}This paper proposes a novel neural architecture to {\color{black}achieve both CF prevention and knowledge transfer, which also deals with the CF problem with BERT fine-tuning.} The proposed system is called \textbf{CTR} (\textit{Capsules and Transfer Routing for continual learning}). %\textit{A}chieving \textit{F}orgetting avoidance and \textit{K}nowledge transfer in continual learning). 
CTR inserts a continual learning plug-in (CL-plugin) module in two locations in BERT. % \zixuan{this actually happen in each transformer layers in BERT}. 
With the pair of CL-plugin modules added to BERT, we no longer need to fine-tune BERT for each task, which causes CF in BERT, and yet we can achieve the power of BERT fine-tuning. CTR has some similarity to Adapter-BERT~\cite{Houlsby2019Parameter}, % \zixuan{the next few sentences introduce Adapter, but seems we still have not said what is exactly is the similarity between adapter and CL-plugin} 
{\color{black}which adds adapters in BERT} for parameter efficient transfer learning such that different end tasks can have their separate adapters (which are very small in size) to adapt BERT for individual end tasks and to transfer the knowledge from BERT to the end tasks. Then, there is no need to employ a separate BERT and fine-tuning it for each task, which is extremely parameter inefficient if many tasks need to be learned. An adapter is a simple 2-layer fully-connected network % \zixuan{inserted in each layer of BERT} 
for adapting BERT to a specific end task. A CL-plugin is very different from an adapter. We do not use a pair of CL-plugin modules to adapt BERT for each task. Instead, CTR learns all tasks using only one pair of CL-plugin modules inserted into BERT. A CL-plugin is a full CL network that can leverage a pre-trained model and deal with both CF and knowledge transfer. Specifically, it uses a \textit{capsule}~\cite{hinton2011transforming} to represent each task and a proposed \textit{transfer routing} algorithm to identify and transfer knowledge across tasks to achieve improved accuracy. It further learns and uses task masks to protect task-specific knowledge to avoid forgetting. %Thus CL-plugins represent a full CL system that can make use of a pre-trained model. 
Empirical evaluations show that CTR outperforms strong baselines. Ablation experiments have also been conducted to study where to insert the CL-plugin module in BERT in order to achieve the best performance (see Sec.~\ref{sec:results}).  
}

\vspace{-2mm}
\section{Related Work}
\label{Sectionrelated.work}
% \zixuan{Section 2 are copied from ECML paper}
% As we are not aware of any reported work that uses BERT fine-tuning and capsules in CL, here we discuss the related work in CL in general with regard to overcoming CF and knowledge transfer. % and then some specific CL approaches for solving the problems that use in our experiments. \hu{$<-$remove?}

\textbf{Catastrophic Forgetting:}
Existing work in CL mainly focused on overcoming CF using the following approaches. 
(1) \textit{Regularization-based approaches,} such as those in~\cite{Kirkpatrick2017overcoming,DBLP:conf/nips/LeeKJHZ17,Seff2017continual,zenke2017continual}, add a regularization in the loss to consolidate weights for previous tasks when learning a new task.
(2) \textit{Replay-based approaches}, such as those in~\cite{Rebuffi2017,Lopez2017gradient,Chaudhry2019ICLR,wang2020efficient}, retain some training data of old tasks and use them in learning a new task. 
The methods in~\cite{Shin2017continual,Kamra2017deep,Rostami2019ijcai,He2018overcoming} learn data generators and generate old task data for learning a new task.
(3) \textit{Parameter isolation-based approaches,} such as those in \cite{Serra2018overcoming,ke2020mixed,Mallya2017packnet,fernando2017pathnet}, allocate model parameters dedicated to different tasks % the parts of the network that are important to the previous tasks 
and mask them out when learning a new task.
(4) \textit{Gradient projection-based approaches}~\cite{zeng2019continuous} ensure the gradient updates occur only in the orthogonal direction to the input of old tasks and thus will not affect old tasks. 
 % followed the Generative Adversarial Networks (GANs) framework \cite{Goodfellow2016NIPSTutorial} to produce data generators for previous tasks, and learn parameters that fit a mixed set of real data of the new task and replayed/generated data of previous tasks. 
Some recent papers used pre-trained models~\cite{hu2021continual,ke2021Classic,ke2021adapting} and learn one class per task~\cite{hu2021continual}.
Tackling CF only deals with model deterioration. These methods perform worse than learning each task separately. An empirical study of the cause of CF and the impact of task similarity on CF was done in~\cite{ramasesh2021anatomy}. 

Some NLP applications have also dealt with CF. For example, CL models have been proposed for sentiment analysis~\cite{ke2021Classic,ke2021adapting,DBLP:conf/dasfaa/LvWLCZ19,qin2020using}, dialogue slot filling~\cite{shen-etal-2019-progressive}, 
language modeling~\cite{sun2020lamol,chuang2020lifelong}, language learning~\cite{li2019compositional}, sentence embedding~\cite{liu2019continual}, machine translation~\cite{khayrallah2018regularized}, cross-lingual modeling~\cite{liu2020exploring}, and question answering~\cite{greco2019psycholinguistics}. A dialogue CL dataset is also reported in \cite{madotto2020continual}. 
% More details can be found in the survey~\cite{biesialska2020continual}. % We focus on Task-CL of supervised NLP classification tasks.

\textbf{Knowledge Transfer:}
% The proposed CTR can perform knowledge transfer to improve the performance of both the new and old tasks as we focus on learning similar natural language tasks, which tend to have shared knowledge. 
Ideally, learning from a sequence of tasks should \textit{also} allow multiple tasks to support each other via knowledge transfer.
CAT~\cite{ke2020mixed} (a Task-CL system) works on a mixed sequence of similar and dissimilar tasks and can transfer knowledge among similar tasks detected automatically. 
Progressive Network~\cite{DBLP:journals/corr/RusuRDSKKPH16} does forward transfer but it is for class continual learning (Class-CL). 
% \cite{ke2020mixed} proposed to learn a mixed sequence of tasks, where similar tasks can transfer knowledge to each other.
% Given the popularity of BERT, we notice the difficulty of naive fine-tuning BERT for CL in NLP applications (see Section \ref{sec:results}), let alone for knowledge transfer.
%specific CL problem in the NLP setting. 

% Since we use two sentiment analysis related problems to conduct our experiments due to the availability of data with many tasks, here we discuss their related work. \hu{maybe not mention in related work but just in experiments?}
% The two sentiment analysis problems that we use are document sentiment classification (DSC) and aspect sentiment classification (ASC). Although some research about CL on DSC has been done, we have not found any existing work on CL for ASC tasks. 

Knowledge transfer in this paper is closely related to \textit{lifelong learning} (LL), which aims to improve the new/last task learning without handling CF~\cite{Silver2013,ruvolo2013ella,chen2018lifelong}. 
In the NLP area, NELL~\cite{carlson2010toward} performs LL information extraction, and several other papers worked on lifelong document sentiment classification (DSC) and aspect sentiment classification (ASC). 
%\hu{maybe remove some non-NN citations later depending on total length.}
\cite{DBLP:conf/acl/ChenM015} and~\cite{hao2019forward} proposed two Naive Bayesian methods to help improve the new task learning.
\cite{xia2017distantly} proposed a LL approach based on voting. \cite{ShuXuLiu2017} used LL for aspect extraction. \cite{qin2020using} and \cite{shuai2018lifelong} used neural networks for DSC and ASC, respectively. {\color{black}Several papers also studied lifelong topic modeling~\cite{chen2018lifelong,gupta2020neural}.} However, all these works do not deal with CF.

{\color{black}SRK~\cite{DBLP:conf/dasfaa/LvWLCZ19} and KAN \cite{ke2020continual} try to deal with both CF and knowledge transfer in continual sentiment classification. However, they have two critical weaknesses: (i) Their RNN architectures cannot use plug-in or adapter modules to tune BERT, which significantly limits their power. (ii) Since they were mainly designed for knowledge transfer, they suffer from serious CF (see Sec.~\ref{sec:results}).  B-CL~\cite{ke2021adapting} uses the adapter idea~\cite{Houlsby2019Parameter} to adapt BERT for sentiment analysis tasks, which are similar to each other. However, since its mechanism of \textit{dynamic routing} for knowledge transfer is very week, its knowledge transfer ability is markedly poorer than CTR (see Sec.~\ref{sec:results}).  CLASSIC~\cite{ke2021Classic} is another recent work on continual learning for knowledge transfer, but its CL setting is \textit{domain continual learning}. Its knowledge transfer method is based on contrastive learning.}
%but they improve only the last task and do not deal with CF and do not use BERT. 

% But, none of the systems uses BERT, which presents a unique challenge to CL as discussed in Section~\ref{Sectionintro}. 
% We will see in the experiment section that the existing CL systems for dealing with DSC are weaker than CTR.  

% \hu{are we the first for capnet on CL?}
% \hu{consider to make CapsNet another bolded title? since it's not very part of knowledge transfer but a base network we use}

% \textbf{Capsule Networks.} CapsuleNet has been used in several NLP applications \cite{DBLP:conf/acl/ChenQ19,DBLP:conf/acl/ZhaoPECY19}. This work adopt capsules \cite{hinton2011transforming} for continual learning. The capsules in our work also have a new routing algorithm, which is instrumental for knowledge transfer.

{\color{black}AdapterFusion~\cite{pfeiffer2020adapterfusion} used adapters proposed in~\cite{Houlsby2019Parameter}. It proposes a two-stage method to learn a set of tasks. In the first stage, it learns one adapter for each task independently using the task’s training data. In the second stage, it uses the training data again to learn a good composition of the learned adapters in the first stage to produce the final model for all tasks. AdapterFusion basically tries to improve multi-task learning. It is not for continual learning and thus has no CF. As explained in Sec.~\ref{sec.intro}, the CL-plugin concept in CTR is different from that of adapters for adapting BERT for each task. CL-plugins are continual learning systems that make use of a pre-trained model. }

\section{CTR Architecture}
\label{Sectionpreliminary}
\begin{figure}[t]
\centering
\includegraphics[width=0.8\columnwidth]{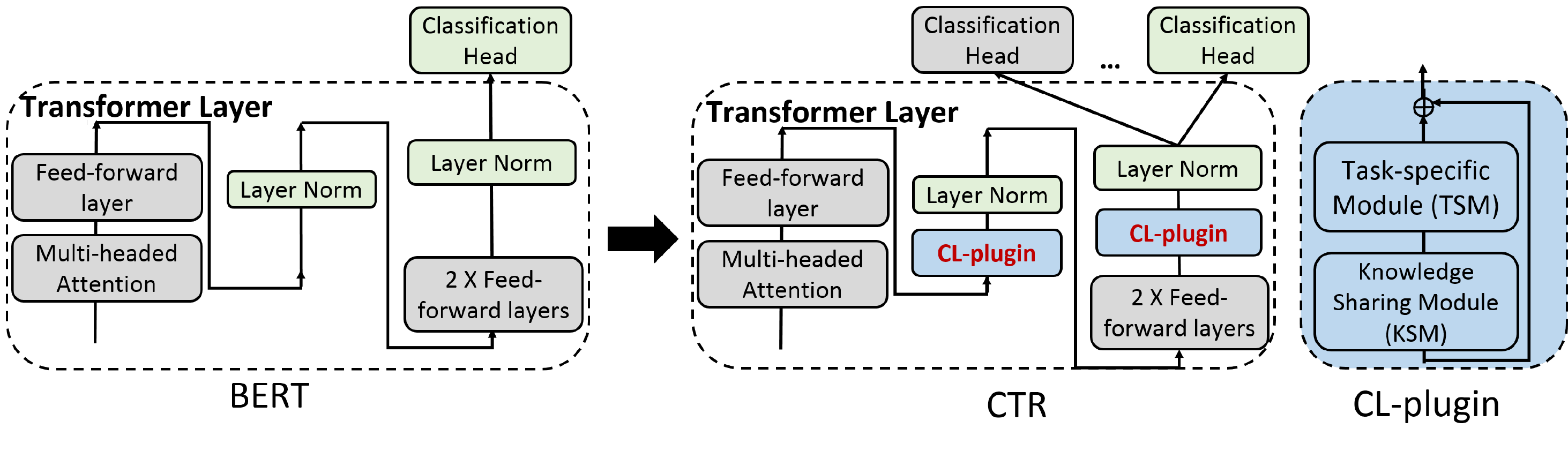}
\caption{
% \hu{you can make rays out of the lower CL-plugin, too.} 
% \hu{heard from Prof and Zixuan's suggestion: what is a good name for the whole pic? Adaformer, CLformer, Transformer-CL? any other ideas?} 
% Adapter-BERT in \cite{DBLP:conf/icml/HoulsbyGJMLGAG19} and our BERT-based continual learning (BACK). 
{\color{black}Architecture of BERT \textbf{(left)} and  
%Only the adapter (yellow boxes) and layer norm (green boxes) layers are trainable. The other modules (grey boxes) are frozen.
the proposed system CTR \textbf{(right)}, which inserts two CL-plugins in BERT. Each CL-plugin module (\textbf{far right})} has two sub-modules and a skip connection: knowledge sharing sub-module (KSM) and task-specific sub-module (TSM). % Each of these modules has a skip-connection.
% \hu{change the color in this figure to make a clear difference than NAACL submission?}
% \ls{modify this figure to make a difference to naacl} \bing{make the two figures horizontal rather than vertical to save space}
}
\label{overview_adapter}
\vspace{-4mm}
\end{figure}

{\color{black}This section describes the general architecture of CTR. The details about its key component CL-plugin is presented in the next section. % introduces BERT, Adapter-BERT and Capsule Network as they are used in our model. 
% \hu{the following bolded title already covers ``Adapting BERT for CL'', change it to Off-the-shelf BERT Fine-tuning for Continual Learning or Naive BERT Fine-tuning for Continual Learning ? }
% \textbf{BERT Fine-Tuning for CL.} %\bing{I suggest that we move the first part to the introduction and only leave the intro to the Adapter here as BERT is well known it is too late to say the issue about BERT not suitable for CL. We have to say it in the intro and then no need to repeat it here. We also need to say why BERT is not suitable to continual learning and why Adapter is suitable.)} 
Due to its good performance, BERT \cite{DBLP:conf/naacl/DevlinCLT19} and its transformer \cite{vaswani2017attention} architecture are used as the base in our model CTR. Since BERT fine-tuning is prone to CF (Sec.~\ref{sec.intro}), we propose the CL plug-in idea, which is inspired by Adapter-BERT~\cite{Houlsby2019Parameter}. 
CL-plugin is a full continual learning module designed to interact with a pre-trained model, in our case, BERT. 
% We would like BERT to be effectively used in continual learning, especially for aspect-based sentiment classification.
%\textbf{From Adapter Mechanism to Continual Adapter.} Our proposed model is based on 
%can be an ideal fit for
% \hu{an ideal fit for?} continual learning. 

\textbf{Inserting CL-plugins in BERT.} 
A commonly used method of leveraging a pre-trained model is to add the end task module on top of the pre-trained model. However, as explained in Sec.~\ref{sec.intro}, fine-tuning the pre-trained model can cause serious CF for CL. The CL system PCL~\cite{hu2021continual}, which uses this approach, has the pre-trained model frozen {\color{black}to avoid forgetting}. But as we will see in Sec.~\ref{sec:results}, this is not the best choice. CTR inserts the proposed CL-plugin in two locations in BERT, i.e., in each transformer layer of BERT. We will also see in Sec.~\ref{sec:results} that inserting only one CL-plugin in one location is sub-optimal. 
%It inserts a 2-layer fully connected network (adapter) in each transformer layer of BERT (see Figure~\ref{overview_adapter}(A)). During fine-tuning, only the weights from adapters (yellow boxes) and normalization layers (green boxes) are trainable. All other BERT parameters (grey boxes) are frozen.
% \cite{DBLP:conf/icml/HoulsbyGJMLGAG19} shows Adapter-BERT achieves similar performances to fine-tuned BERT in NLP tasks with less trainable weights. %in the non-CL setting. 
Figure~\ref{overview_adapter} gives the CTR architecture and we can see the two CL-plugins are added into BERT. In learning, only the two CL-plugins and the classification heads are trained. The components of the original pre-trained BERT are fixed}.  %, which we will discuss in the next section. %, freezing weights in transformers and only train adapters (instead of training multiple layers of transformers as a whole) can be beneficial regarding performance (see Section~\ref{sec:results}), number of trainable parameters and more importantly, easy incorporation for existing models with continual learning capability.
%  By \textit{only} training the adapter, \textit{without} any changing of the original BERT parameters, the model can achieve similar performance as conventional BERT fine-tuning. This is attempting because it can largely reduce the training parameters. More importantly, this also makes it possible that one may enable BERT working in continual learning scenario by only modifying the adapter. \hu{maybe shorten this part?}

\begin{figure*}[t]
\centering
\includegraphics[width=4.8in]{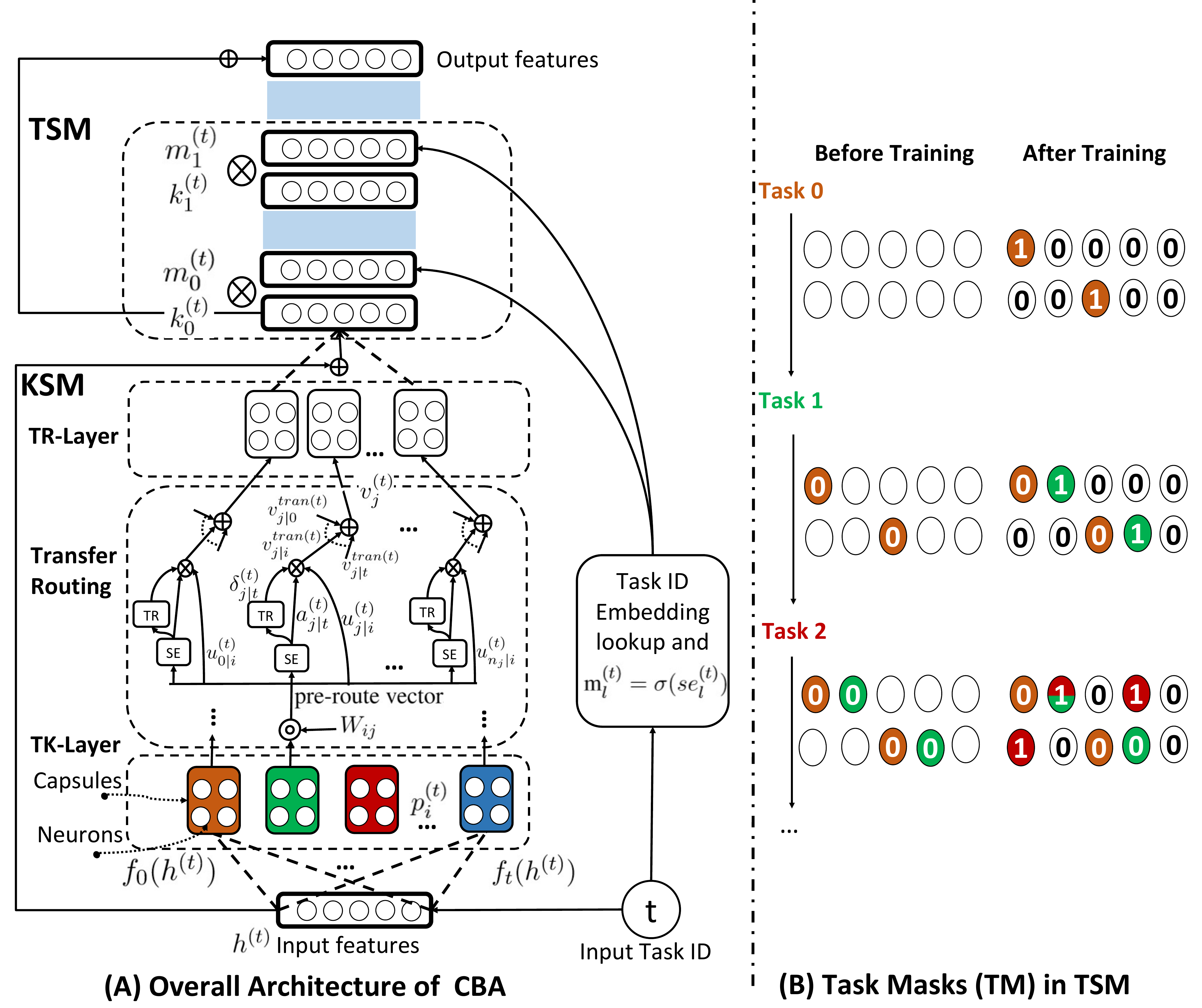}
\caption{%\hu{two types of red are hard to tell the difference, esp. in BW printing?}
% \hu{how about task embedding lookup in the figure?}
% \hu{don't know y this fig seems vertically strenched.}
\textbf{(A)} CL-plugin Architecture.
% : the skip-connection is not shown for clarity. 
\textbf{(B)} Illustration of task masking. {Cells/neurons in brown, green and red are respectively used by tasks 0, 1 and 2. Neurons with two colors are used by two tasks} % \ls{(B) task 2: do 1 and 0 use the same color (brick red)? if color is related to a task id, can we make "task id" highlighted by the corresponding color.}% each previous task representation compare with current task representation via Eq. (\ref{eq.transcl.relu})$\to$(\ref{eq.transcl.multi}) and results in transferable representation; Dynamic routing: weighted sum on all the task representations have been seen so far (previous and current task) via Eq. Eq.~(\ref{eq:task_shared_features})$\to$(\ref{eq.squash}), resulting in clustering representations; Task masking: a (learnable) task mask is applied after the activation function to \textit{selectively} activate a neuron (or feature). Some notes about (B) are: the two rows of each task corresponds to $k_0^{(t)}$ and $k_1^{(t)}$ in TSM. In the cells 
% \hu{re-describe what the matrix is about}
% before training, those solid cells are the neurons to be protected (masked) and those cells without a number are free neurons (not used). In the cells after training, those solids cells show neurons that have been used. In the cells Masks, those with 0’s are the neurons to be protected (masked); those cells with 1’s show neurons that are important for the current task, which are used as a mask for the future. Note that those cells with multiple times 1 indicate that they are shared by more than one task. % Their gradients are blocked in backward propagation since they are important neurons for previous tasks. %The rest of the cells or neurons are not important for the task. 
% Those 0 cells without a color are not used by any task.
}
\vspace{-3mm}
\label{overview}
\end{figure*}

\textbf{Continual learning plug-in (CL-plugin).} CL-plugin employs a capsule network (CapsNet)~\cite{hinton2011transforming,sabour2017dynamic} like architecture. In the classic neural network, a neuron outputs a scalar, real-valued activation as a feature detector. CapsNets replaces that with a vector-output capsule to preserve additional information. A simple CapsNet consists of two capsule layers. The first layer stores low-level feature maps, and the second layer generates the classification probability with each capsule corresponding to one class. CapsNet uses a \textit{dynamic routing} algorithm to make each lower-level capsule to send its output to a similar (or ``agreed'', {\color{black}computed by dot product)} 
higher-level capsule. This property can already be used to group similar tasks and their shareable
features to produce a CL system (see the ablation study in Sec.~\ref{sec:results}). 
One of the key ideas of CL-plugin (see Figure~\ref{overview}(A)) is
a \textit{transfer capsule layer} with a new \textit{transfer routing} algorithm to explicitly identify transferable features/knowledge from previous tasks to transfer to the new task.
Additionally, transfer routing avoids the need for hyper-parameter tuning on the number of iterations of dynamic routing~\cite{sabour2017dynamic}
% does not need multiple iterations 
to update the agreements.

% CapsNet is able to adaptively decide the information transferred between layers by using dynamic routing. It is also able to omit the disagreed output of lower layers in the dynamic routing algorithm, which encourages learning of generalizable features. Therefore, CapsNet meets our needs of generating general and task sharing knowledge (details in Section \ref{sec:task_sharing}). % To our knowledge, our proposed model is the first attempt to exploit the power of CapsNet under the continual learning framework for ASC tasks.

% Note that the proposed CTR does not adopt the whole capsule network as we are only interested in the capsule layers and dynamic routing instead of the max-margin loss and the classifier. 

% \textbf{Neuron Masking. ?}
% \hu{the masking part is too long in the next section. bring common knowledge to here ?}.

\section{Continual Learning Plug-in (CL-plugin)}
% \vspace{-1mm}

The architecture of our \textit{continual learning plug-in} (CL-plugin) is shown in Figure~\ref{overview}(A). CL-plugin takes two inputs: (1) hidden states $h^{(t)}$ from the feed-forward layer inside a transformer layer and (2) task ID $t$, which is required by task continual learning (Task-CL).
The outputs are hidden states with features suitable for the $t$-th task for classification. Inside CL-plugin, there are two modules: (1) \textit{knowledge sharing module} (KSM) for identifying and transferring the shareable knowledge from similar previous tasks to the new task $t$, and (2) \textit{task specific module} (TSM) for learning task specific neurons and their masks (which can protect the neurons from being updated by future tasks to deal with CF). Since TSM is differentiable, the {whole system CTR can be trained end-to-end.} % We detail each module below. 

% TSM allocates neurons conservatively so that more unused neurons can be dedicated to shareable knowledge from KSM for the new task.
%\hu{do we need to explain why KSM goes first and TSM second because existing research show that lower layer of DL models tend to learn more general features?}

%learned from both the new and old tasks. 

% KSM leverages capsule layers (see below) and dynamic routing to group similar tasks and the shareable knowledge, whereas

%are then fed into a dynamic capsule layer (KCL) to exploit shareable knowledge across all previous and current tasks by the dynamic routing algorithm.
%We can obtain the task sharing features after the combination. 

% TSM takes advantage of task mask (TM) to protect neurons for a particular task and leave other neurons free. Those free neurons are later used by TSM for a new task.

%new tasks with KSM\zixuan{TSM} also leveraging 
 % \hu{better now?}
%A task specific module then takes as input the task sharing features and task ID to learn task specific knowledge (TSK) and also a binary task mask (TM) for each layer in TSK. 

%A TM indicates the neurons that are important/useful for the new task in the layer and is used to protect the learned knowledge from being overwrite by a later task\hu{check if this statement is accurate.}.

% More specifically, the task sharing knowledge among tasks is aggregated and summarized to allow knowledge transfer, whereas task specific knowledge is identified to avoid being catastrophically forgotten when new tasks come in.

\subsection{Knowledge Sharing Module (KSM)}
\label{sec:task_sharing}
\vspace{-0.5mm}
% \zixuan{Semancitc capsule -> Task capsule layer, each capsule represent feature of each task}}

% \zixuan{KSL -> clustering similar task, similar tend to be updated: backward transfer (may wrong for mixed sequence of task)}

% \zixuan{TSM -> we did not limit share with previous task but block the updating}

KSM achieves knowledge transfer among similar tasks via a \textit{task capsule layer} (TK-Layer), a \textit{transfer capsule layer} (TR-Layer), and a \textit{transfer routing} mechanism.

% Dynamic routing and a novel transfer routing are employed.

%and two capsule layers are employed to achieve this. 
% \hu{is layer more accurate? I think capnet arch includes all those cnn layers to classifiers.} 
% KSM has two types of capsule layers and leverages a dynamic routing algorithm to selectively combine knowledge from the old and new tasks.

\subsubsection{Task Capsule Layer (TK-Layer)} %\ls{TK-Layer makes me think task continual learning. I feel CL is overly used.}
Each capsule in the TK-Layer represents a task, and it prepares the low-level features derived from each task (Figure~\ref{overview}(A)). As such, a capsule is added to the TK-Layer for each new task. This incremental growth is efficient and easy because these capsules are discrete and do not share parameters.
% \hu{not sure why KCL is related to TCL's easy and efficient?} 
Also, each capsule is simply a 2-layer fully connected network with a small number of parameters. %(\bing{?? something wrong with this sentence}).
Let $h^{(t)}\in\mathbb{R}^{d_t\times d_e}$ be the input of CL-plugin, where $d_t$ is the number of tokens, $d_e$ the number of dimensions, and $t$ is the current task. % \hu{introducing t-1 separately from t doesn't have help on explanation to me but a bit verbose here? Just say ``Assume the CL system reaches to learn the $t$-th task from a sequence of tasks'' before line 502 in latex Let ...}
% Let the number of tasks learned so far be $t-1$ (before learning the new/current task, the $t$-th task). 
In the TK-Layer, we have one capsule for each task. Assume we have learned $t$ tasks so far. % $t$ different capsules representing all past $t-1$ learned tasks as well as the new $t$-th task. 
%(\bing{?? is this fixed upfront?}). 
The capsule for the $i$-th ($i\leq t$) task is
\begin{equation}
\label{eq:task_specific}
p^{(t)}_i = f_i(h^{(t)}),
\end{equation}
where $f_i(\cdot)=\text{MLP}_i(\cdot)$ denotes a 2-layer fully-connected network.

% \zixuan{\textbf{Task Routing.} Since we have different MLPs for different tasks and we need compute shared knowledge on those task representation. Therefore, task routing is intuitive and simple send each task representation to next layer.}

% \zixuan{

\subsubsection{Transfer Routing and Transfer Capsule Layer}
% \zixuan{DynCL has been commented out}

% \zixuan{Original TR-Layer has been commented out}
% \bing{We need to have a clear separation of transfer routing and transfer capsules.}
% \zixuan{
% {\color{blue}Given the task capsule from TK-Layer, TR-Layer aims to extract the transferable proportions and send them to the higher-level layer, the \textit{transfer capsules}. 

{Each capsule in the \textit{transfer capsule layer} (TR-Layer) represents the transferable representation extracted from TK-Layer. As shown in Figure 2(A), \textit{transfer routing} between the lower-level capsules in TK-Layer and high-level capsules in TR-Layer has three components: {\textit{pre-route vector generator}} (\textit{PVG}), \textit{similarity estimator (SE)} and \textit{task router (TR)}. Given the task capsules in the TK-Layer, we first transform the feature through a trainable {weight matrix}. We call the output of this transformation the \textit{pre-route vector}. Each SE estimates the similarity between a previous task and the current task using the pre-route vector, resulting in a similarity score for each higher-level capsule. Additionally, each SE is augmented by a TR module, a differentiable task router acting as a gate. This router estimates a binary signal that decides whether to connect or disconnect the current route between the two consecutive capsule layers (i.e. TK-Layer and TR-Layer in CL-plugin). The binary signal estimated by TR can be seen as 
% \hu{the gumbel-softmax trick makes non-differiable differentiable, it's a trick (detail) as a solution, I favor just say hard-attention or sampling on high-level here (and later GS trick to make it differentiable). Keeping both seems a conflict.) }
a differentiable binary attention. % {\color{red}gate} \zixuan{if use "gate", seems repeated to line 198} coefficient. 
% Conceptually, each route can be seen as a double-attention(soft similarity and hard gate) mechanism
Conceptually, each SE and TR pair together learns the connectivity between capsules in a stochastic and differentiable manner, which can be seen as a task similarity-based connectivity search mechanism. This transfer routing identifies the shared features/knowledge from multiple task capsules and helps knowledge transfer across similar tasks. 
Next, we discuss the {\textit{pre-route vector generator}}, \textit{similarity estimator} and \textit{task router}.

{\textbf{Pre-route Vector Generator (PVG).}}
We first turns each transfer capsule $p_{i}^{(t)}$ into a pre-route {vector}, % $u_{j|i}^{(t)}$ 
\begin{equation}
% \vspace{-1.5mm}
\label{eq:pre_shared_features}
u^{(t)}_{j|i} = W_{ij}p_{i}^{(t)},
% \vspace{-3mm}
\end{equation}
where $W_{ij}\in\mathbb{R}^{d_s\times d_k}$ is the weight matrix, $d_s$ and $d_k$ are the dimensions of task capsule $i$ (also representing a task) and transfer capsule $j$, and $t$ is the current task. The number of transfer capsules $n_j$ is a hyperparameter detailed in Sec.~\ref{Sectionexperiments}.

\textbf{Similarity Estimator (SE).}
Since tasks $i$ and $t$ are different, it is crucial to determine what in task $i$'s representation is transferable. Inspired by \cite{DBLP:conf/acl/LiX18}, we use a convolution layer and activation units to compare task $i$ with task $t$ to determine the transferable proportion from the previous task $i$. In SE, we compute the task similarity as follows: %  \zixuan{fixed a typo mentioned in rebuttal}: % (\bing{?? what is j?}):
\begin{equation}
\label{eq.transcl.cur}
    q_{j|t}^{(t)} = \text{MaxPool}(\text{Relu}(u^{(t)}_{j|t}*W_q+b_q)),
    \vspace{-1mm}
\end{equation}
\begin{equation}
\label{eq.transcl.sim}
% \vspace{-3mm}
    a_{j|i}^{(t)} = \text{MaxPool}(\text{Relu}(u^{(t)}_{j|i}* W_a + f_a(q_{j|t}^{(t)}) +b_a)),
    % \vspace{-3mm}
\end{equation}
where $b_a,b_q\in\mathbb{R}$ are the bias, $W_a,W_q\in\mathbb{R}^{d_e\times d_w}$ 
% \hu{avoid variables of double chars, that means multiplication; could we find a single variable?}
are convolutions filters and $d_w$ is the windows size. We extract important features from the current task representation $u_{j|t}^{(t)}$ via the convolution network in Eq. (\ref{eq.transcl.cur}). The MaxPool helps remove the insignificant features to generate a fixed-size vector with the size equal to the number of filters $n_{w}$. Similarly, we extract important features from the previous task $i$'s representation $u_{j|i}^{(t)}$. Using the important features for the current and previous tasks, we compute a similarity score between them in Eq.~(\ref{eq.transcl.sim}) with ReLU activation. Note $f_a$ is a 1-layer fully-connected network to match the dimensions. As a result, $a_{j|i}^{(t)}$ indicates how similar the representation of the $i$-th task is to the current task $t$. For those tasks with a very low $a_{j|i}^{(t)}$, their representations are less similar to the current task and thus has little transferable knowledge. 

\textbf{Task Router (TR).} %\hu{again TR seems a confusing acronym}}
TR controls which previous task representation should flow to the next layer with the goal of letting only the transferable information to flow. Given the similarity $a_{j|i}^{(t)}$, TR estimates a binary decision signal $\delta_{j|t}^{(t)} \in \{\texttt{\textbf{0}:disconnect, \textbf{1}:connect}\}$. We first apply a convolution layer with 2 output channels and $1\times1$ kernel size to generate un-normalized decision value. To estimate the binary decision, we need to generate a decision chosen from the set of two mutually exclusive and exhaustive events (disconnect and connect). In our work, we adopt the Gumbel-Softmax \cite{DBLP:conf/iclr/JangGP17} to help make the TR gate %{\color{red}gate} \zixuan{sampling} 
differentiable. 
\begin{equation}
\label{eq.transcl.gate}
    \delta_{j|i}^{(t)} = \text{Gumbel\_softmax}(a_{j|i}^{(t)}* W_{\delta}+b_{\delta}).
    % \vspace{-3mm}
\end{equation}
Given the similarity $a_{j|t}^{(t)}$, binary decision $\delta_{j|t}^{(t)}$ and the pre-route vector $u_{j|t}^{(t)}$, we compute the transferable representation $v_{j|i}^{\text{tran}(t)}$ and final output $v_{j}^{(t)}$ as follows: 
% \bing{reviewer 1 says v -> u?}\zixuan{changed the pre-route vector}:
\begin{equation}
\label{eq.transcl.tran}
    v_{j|i}^{\text{tran}(t)} = {a_{j|i}^{(t)} \otimes u_{j|i}^{(t)}},~~~~~~
        v_{j}^{(t)} = \sum^{n+1}_{i=1\atop\delta_{ij}^{(t)}=1}v_{j|i}^{\text{tran}(t)}.
%    \vspace{-1mm}
\end{equation}
%\begin{equation}
%\label{eq.transcl.final}
%    v_{j}^{(t)} = \sum^{n+1}_{i=1\atop\delta_{ij}^{(t)}=1}v_{j|i}^{\text{tran}(t)}.
%    \vspace{-3mm}
% \end{equation}
This makes sure only task capsules for tasks that are salient or similar to the new task are used, and the others task capsules are ignored (and thus protected) to learn more general shareable knowledge. As many NLP applications have similar tasks, such learning of task-sharing features can be very important. Note that in backpropagation, only the similar tasks with connected gate ($\delta_{ij}^{(t)}$=1) are updated, encouraging backward knowledge transfer of similar tasks.
}

\subsection{Task Specific Module (TSM)} 
We now discuss how to preserve task-specific knowledge of previous tasks to prevent forgetting (CF). % and leave more neurons for knowledge sharing.
%Besides knowledge sharing, we also need to capture the task-specific knowledge. 
%In learning the task specific knowledge, we want to 
%We want to ensure the task specific knowledge of old tasks that cannot be shared with the new task is not changed by the new task; otherwise sever catastrophic forgetting (CF) occurs.
 % to restrict the modification of neurons and keep such masks consistent throughout the entire CL process. 
To achieve this, we use task masks (Figure~\ref{overview}(B)). Specifically, we first detect the neurons used by each old task and then block off or mask out all the \textit{used} neurons when learning a new task. 

The task-specific module consists of %layers with 
%The key component in task specific module is the 
%task specific knowledge, which can be 
differentiable layers (a 2-layer fully-connected network is used).
Each layer's output is further applied with a task mask to indicate which neurons should be protected for that task to overcome CF and forbids gradient updates for those neurons during backpropagation for a new task. 
Those tasks with overlapping masks indicate some parameter sharing. %\hu{to encourage knowledge sharing}.
%\hu{check if we can claim this or not} 
Due to KSM, the features flowing into those overlapping neurons enable the related old tasks to also improve in learning the new task. {\color{black}Here we borrow the hard attention idea in \cite{Serra2018overcoming} and leverage the task ID embedding to train the task mask. Further details can be found in Supplementary.}

\textbf{Illustration.} The task masking process is illustrated in Figure~\ref{overview}(B), which shows the learning process of three tasks. Before training, those solid cells with a 0 are the neurons that have been used by some previous tasks and should be protected (masked). Those empty cells are free neurons (not used). After training, those solid cells with a 1 are neurons that are important for the current task, which will be used as masks in the future. Those solid cells with a 0 are masked as they are important for previous tasks. Those non-solid cells with a 0 are neurons that are not used so far.  % Note that those cells with multiple times 1 \bing{no multiple 1's} indicate that they are shared by more than one task.

Let us walk through the learning process of the three tasks. After training task 0, we obtain its useful neurons indicated by the 1 entries. Before training task 1, those useful neurons for task 0 are first masked (those previous 1's entries are turned to 0's). After training task 1, two neurons with 1 are used by the task. When task 2 arrives, all used neurons by tasks 0 and 1 are masked before training, i.e., their entries are set to 0. After training task 2, we see that tasks 2 and 1 have a shared neuron (the cell with two colors, red and green), which is used by both of tasks. % We mark the shared neuron 1, although it has been used by task 0.

\section{Experiments}
\label{Sectionexperiments}
We evaluate CTR using three applications. We follow the continual learning (CL) evaluation method given in~\cite{DBLP:journals/corr/abs-1909-08383}.
% \hu{not sure if ICML reviewers already know these well (seems totally not a problem for naacl reviewers in sentiment).}
CTR first learns a sequence of tasks. After a task is trained, the training data of the task is no longer accessible. After all tasks are learned, their task models are tested using their respective test sets. %\hu{do we need to breakdown clearly how to evaluate back/forward transfer and forgetting here?} \hu{is this sentence obvious (in CL)?}
In training each task, we use its validation set to decide when to stop training. 

% \zixuan{BERT(frozen) added, Adapter-BERT added, B-CL in ablation study}

% \hu{I think it's harder for reviewers to understand CL evaluation at the beginning. make some research questions you want to focus by priority and recall them CTR in result analysis. this should be consistent with your arguments in intro, tech part.}
% \zixuan{Do we need this? say RQ1. In continual learning setting, which models results in forgetting, how much and can CTR adreess this; RQ2. How is the effectiveness of backward and forward knowledge transfer of CTR; RQ3: Is each part of the CTR contributes to its final performance }

\subsection{Three Applications and Their Datasets}

The first two applications (and datasets) are used to show the knowledge transferability of CTR because their tasks are similar and have shared knowledge. Catastrophic forgetting (CF) is not a major concern for them. The third application (and dataset) is mainly used to test CTR's ability to overcome CF as its tasks are very different and have little shared knowledge to transfer.  

\textbf{1. Document Sentiment Classification (DSC).} This application is to classify each full product review into one of the two opinion classes (\textit{positive} and \textit{negative}). The training data of each task consists of
% a set of positive and negative 
reviews of a particular type of product. %\footnote{The CL/Task-CL is important for opinion data due to privacy reasons as many clients/users want their private opinion data deleted after use. In such cases, if we want to improve accuracy without breaching confidentiality, Task-CL is a suitable solution.}
% In the chatbot context, users do not want their data uploaded to a central server. 
% In such applications, if we want to improve some task for each user/client without breaching confidentiality, CL is a suitable solution.
%DSC on different tasks could be similar and thus have sharable knowledge as sentiment expressions % used on different products are similar across products. 
We adopt the text classification formulation in \cite{DBLP:conf/naacl/DevlinCLT19}, where a \texttt{[CLS]} token is used 
% added at the beginning of the review and the sentiment polarity is
to predict the opinion polarity.
% predicted on top of this special token.

We employ a set of 10 DSC datasets (reviews of 10 products) to produce sequences of tasks. The products are Sports, Toys, Tools, Video, Pet, Musical, Movies, Garden, Offices, and Kindle~\cite{ke2020continual}. Two experiments are conducted: (1) using small data in each task: 100 positive and 100 negative training reviews per task; (2) using the full data in each task: 2500 positive and 2500 negative training reviews per task~\cite{ke2020continual}. (1) is %\zixuan{remove "a"} 
more useful in practice because labeling a large number of examples is very costly. The same validation reviews (250 positive and 250 negative) and the same test reviews (250 positive and 250 negative) are used in both experiments. 

% \footnote{The original data has 5000 training reviews for each task. We did not use all of them because fewer training examples (few-shot) are more realistic in practice and also because BERT is very powerful. If all 5000 reviews are used in training, the accuracy will be very high. CL and knowledge transfer are not needed.} 

% \ls{is 250 positive and 250 negative? we mentioned the label distribution in training but not clearly stated the distribution in testing and validation.}
 
% All existing CL approaches for this application have not used BERT.

\textbf{2. Aspect Sentiment Classification (ASC).} It classifies a review sentence on the aspect-level sentiment (one of \textit{positive}, \textit{negative}, and \textit{neutral}).
% learns a classifier from a set of review sentences about a \textit{product} or \textit{service} (e.g., camera). 
% The model is then used to classify whether a sentence containing a given aspect (a product attribute) expresses a \textit{positive}, \textit{negative}, or \textit{neutral} opinion about the aspect. 
For example, the sentence ``\textit{The picture is great but the sound is lousy}'' about a TV expresses a \textit{positive} opinion about the aspect ``picture'' and a \textit{negative} opinion about the aspect ``sound.'' % \ls{is an example with multi-aspect better (different polarity)? Single-aspect ASC seems similar to DSC.} % ASC is a core problem of sentiment analysis~\cite{liu2015sentiment}.\footnote{Another core problem is \textit{aspect extraction}, e.g., extracting \textit{picture quality} from the above sentence~\cite{liu2015sentiment} and it is not studied in this paper.}
We adopt the ASC formulation in \cite{DBLP:conf/naacl/XuLSY19}, where the aspect term and sentence are concatenated via \texttt{[SEP]} in BERT. The opinion is predicted on top of the \texttt{[CLS]} token.

We employ a set of 19 ASC datasets (review sentences of 19 products) to produce sequences of tasks. Each dataset represents a task. The datasets are from 4 sources: (1) \textbf{HL5Domains} \cite{hu2004mining} with reviews of 5 products; (2) \textbf{Liu3Domains} \cite{liu2015automated} with reviews of 3 products; (3) \textbf{Ding9Domains} \cite{ding2008holistic} with reviews of 9 products; and (4) \textbf{SemEval14} with reviews of 2 products - SemEval 2014 Task 4 for laptop and restaurant. % and restaurant as existing research frequently uses this version. 
For (1), (2) and (3), we use about 10\% of the original data as the validate data, another about 10\% of the original data as the testing data. For (4), we use 150 examples from the training set for validation. To be consistent with existing research \cite{tang-etal-2016-aspect},
sentences with conflict opinions 
% (both positive and negative) 
about a single aspect are not used.
% dropped due to a very small number of examples. 
Statistics of the 19 datasets are given in Table \ref{tab:dataset_main}.
% \footnote{http://qwone.com/~jason/20Newsgroups/}

\textbf{3. Text classification using 20News data}. This dataset \cite{Lang95} has 20 classes and each class has about 1000 documents. The data split for train/validation/test is 1600/200/200. We created 10 tasks, 2 classes per task. Since this is topic-based text classification data, the classes are very different and have little shared knowledge. As mentioned above, this application (and dataset) is mainly used to show CTR's ability to overcome forgetting. 

% We want to achieve two objectives for continual learning of ASC tasks: (1) achieve both forward and backward knowledge transfer across tasks to achieve improved accuracy than without CL, as ASC tasks of different domains are similar and have shared knowledge, and (2) maintain or even improve the performance of the models for previous tasks so that they are not forgotten in learning the new task. 

% \subsection{Experiment Datasets}

\begin{table}[]
\centering
\resizebox{\columnwidth}{!}{
\begin{tabular}{cccccccccccccccccccc}
\specialrule{.2em}{.1em}{.1em}
\multicolumn{1}{c||}{\begin{tabular}[c]{@{}c@{}}Data \\ source\end{tabular}} & \multicolumn{3}{c|}{Liu3domain} & \multicolumn{5}{c|}{HL5domain} & \multicolumn{9}{c|}{Ding9domain} & \multicolumn{2}{c}{SemEval14} \\

\multicolumn{1}{c||}{\begin{tabular}[c]{@{}c@{}}Task/\\ domain\end{tabular}}  &  \rotatebox{90}{Speaker} & \rotatebox{90}{ Router } & \multicolumn{1}{c|}{\rotatebox{90}{Computer}} & \rotatebox{90}{ Nokia6610 } & \rotatebox{90}{ Nikon4300 } & \rotatebox{90}{ Creative } & \rotatebox{90}{ CanonG3 } & \multicolumn{1}{c|}{\rotatebox{90}{ApexAD}} & \rotatebox{90}{ CanonD500 } & \rotatebox{90}{ Canon100 } & \rotatebox{90}{ Diaper } & \rotatebox{90}{ Hitachi } & \rotatebox{90}{ Ipod } & \rotatebox{90}{ Linksys } & \rotatebox{90}{ MicroMP3 } & \rotatebox{90}{ Nokia6600 } & \multicolumn{1}{c|}{\rotatebox{90}{Norton}} & \rotatebox{90}{Restaurant} &  \rotatebox{90}{Laptop}\\

\specialrule{.1em}{.05em}{.05em}
\multicolumn{1}{c||}{Train} & 352 & 245 & \multicolumn{1}{c|}{283} & 271 & 162 & 677 & 228 & \multicolumn{1}{c|}{343} & 118 & 175 & 191 & 212 & 153 & 176 & 484 & 362 & \multicolumn{1}{c|}{194} & 3452 & 2163\\
\multicolumn{1}{c||}{Val.} & 44 & 31 & \multicolumn{1}{c|}{35} & 34 & 20 & 85 & 29 & \multicolumn{1}{c|}{43} & 15 & 22 & 24 & 26 & 19 & 22 & 61 & 45 & \multicolumn{1}{c|}{24} & 150 & 150\\
\multicolumn{1}{c||}{Test} & 44 & 31 & \multicolumn{1}{c|}{36} & 34 & 21 & 85 & 29 & \multicolumn{1}{c|}{43} & 15 & 22 & 24 & 27 & 20 & 23 & 61 & 46 & \multicolumn{1}{c|}{25} & 1120 & 638\\
\specialrule{.1em}{.05em}{.05em}
\end{tabular}
}
\caption{Statistics of datasets for ASC. 
The datasets statistics for DSC and 20News have been described in the text. More detailed data statistics are given in \textit{Supplementary}. % \bing{reformat table to save some space}\zixuan{reformated}
} %\ls{DSC waste too much space}}
\label{tab:dataset_main}
\vspace{-6mm}
\end{table}

\subsection{Baselines}
\label{sec:baselines}

We setup 38 baselines, including both \textit{standalone} and \textit{continual learning} methods. 

\textbf{Multitask learning (MTL}: Results of multitask learning is considered the upper-bound of those of continual learning. Here MTL fine-tunes BERT. 
% for task continual learning (TCL) in different categories and scenarios. The first scenarios is 
% \hu{I'm thinking about a name not using negation: negation always does not know what it really means. Standalone Baselines?}

\textbf{Standalone (SDL) Baselines}: The SDL setting builds a model for each task independently using a separate network. It clearly has no knowledge transfer or forgetting. We have 4 baselines under SDL, \textbf{BERT}, \textbf{BERT (Frozen)}, \textbf{Adapter-BERT} and \textbf{W2V} (word2vec embeddings). For \textbf{BERT}, we use trainable BERT, i.e., fine-tuning.  {\textbf{BERT (Frozen)} uses frozen BERT with a {trainable} CNN text classification network \cite{DBLP:conf/emnlp/Kim14} on top of it.}; \textbf{Adapter-BERT} adapts the BERT as in \cite{Houlsby2019Parameter}, where only the adapter blocks % (2-layer fully connected network) 
are trainable; %, without changing any of BERT parameters;  
\textbf{W2V} uses embeddings 
trained on the Amazon review data in \cite{Xu2018pro} using FastText \cite{grave2018learning}. For ASC,
% \ls{$\leftarrow$ ?? we} 
we adopt the ASC classification network in~\cite{DBLP:conf/acl/LiX18}, which takes aspect term and review sentence as input. For DSC and 20News,
% \ls{$\leftarrow$ ?? we} 
we adopt the classification network in~\cite{DBLP:conf/naacl/DevlinCLT19}.

\textbf{Continual Learning (CL) Baselines}. {\color{black}CL setting includes 4 baselines with % \hu{again, negation could be ambiguous, Naive Fine-tuning?}
\textit{no forgetting handling} (\textbf{NFH}) (corresponding to the 4 standalone baselines), and 25 baselines from 9 state-of-the art \textit{task continual learning} (Task-CL) methods that deal with forgetting (CF). NFH baselines 
learn the tasks one by one with no awareness of forgetting/transfer.
% learn a sequence of tasks incrementally without explicitly tackling forgetting or knowledge transfer. 
% The 4 NFH baselines are \textbf{(5) BERT}, {\textbf{(6) BERT (Frozen)}}, \textbf{(7) Adapter-BERT} and \textbf{(8) W2V}.

The 12 state-of-the-art CL systems are:} % They gives 6 baselines using BERT (Frozen) as features and another 6 baselines using W2V as features. They are unable to use Adapter-BERT or allow BERT to train during learning.  % models for the above three categories, we also consider the state-of-the-art continual learning models. Specifically, they are \textbf{LSTM} \cite{ke2020continual} - a conventional LSTM for classification,
\textbf{KAN} \cite{ke2020continual} and \textbf{SRK} \cite{DBLP:conf/dasfaa/LvWLCZ19} are Task-CL methods %continual learning method proposed to address forgetting and encourage knowledge transfer 
for document sentiment classification. % {\color{red}Their RNN architectures cannot use Adapters. and thus severally limits their power. (see Sec.~\ref{Sectionrelated.work} for more details) }
%It achieves only limited forward transfer. 
The rest were designed initially for image classification. Therefore, we replace their original MLP or CNN image classification network with CNN for text classification \cite{DBLP:conf/emnlp/Kim14}. 
\textbf{HAT} 
\cite{Serra2018overcoming} is one of the best Task-CL methods with almost no forgetting. \textbf{CAT}~\cite{ke2020mixed}
is similar to HAT but can work with a mixed sequence.  %\vspace{-1mm}
\textbf{UCL}
\cite{DBLP:conf/nips/AhnCLM19} is a recent Task-CL method. % using a Bayesian online learning framework. 
\textbf{EWC} \cite{Kirkpatrick2017overcoming} is a popular regularization-based class continual learning (Class-CL) method, which was adapted %Since OWM is a CCL method, but we need a Task-CL method, we adapt it 
for Task-CL by only training on the corresponding head of the specific task ID during training and only considering the corresponding head's prediction during testing.
\textbf{OWM} \cite{zeng2019continuous} is also a Class-CL method, which we also adapt to Task-CL. \textbf{L2} \cite{Kirkpatrick2017overcoming} is a classic regularization based Class-CL method, which we adapt to Task-CL.  \textbf{A-GEM} \cite{Chaudhry2019ICLR} is an efficient version of the replay Task-CL method GEM \cite{Lopez2017gradient}, which penalizes the previous task loss from being increased. \textbf{DER++} \cite{buzzega2020dark} is a recent replay method using distillation to regularize the new task training and it can function as a Task-CL method. {\color{black}\textbf{B-CL}~\cite{ke2021adapting} is similar to CTR but uses dynamic routing for knowledge transfer and it performs markedly worse than CTR. \textbf{LAMOL}~\cite{sun2020lamol} is a pseudo-replay method based on GPT-2.}
% \bing{need a bit more descriptions.}%in the same way.  %Since OWM is a Class-CL method but we need a Task-CL method, we adapt it 
% for Task-CL by only training on the corresponding head of the specific task id during training and only considering the corresponding head's prediction during testing, 
%which slows down learning for weights that are important to the previous tasks.

From the 10 systems, we created {10 baselines} using \textbf{W2V} embeddings with the aspect term added before the sentence so that the Task-CL methods can take both aspect and the review sentence for ASC; {7 baselines} using \textbf{Adapter-BERT} (SRK, KAN and CAT's architectures cannot work with adapters); and {10 baselines} using \textbf{BERT (Frozen)} (which replaces W2V embeddings). The BERT formulation in Sec.~\ref{Sectionpreliminary} naturally takes both aspect and review sentence in the ASC case. 
% For DSC, BERT takes the full reviews. % {\color{red}(?? anything specific for DSC)} %Note that we consider above state-of-the-art continual learning systems in \textbf{W2V} and \textbf{BERT (Frozen)} categories. For W2V, they only takes as input the review sentence since they are originally designed for ASC. For BERT (Frozen), we simply replace the word2vector embedding with frozen BERT and thus takes as input only the review sentence. 

% \subsection{Network and Training Details}
\subsection{Hyper-parameters}
% \ls{what model-size we use to report scores? icml reviewers raised converns on model-size against performance gain}
Unless otherwise stated, % {both DSC and ASC datasets \ls{applications, dataset does not have parameters} 
the same hyper-parameters are used in experiments for ASC, DSC and 20News datasets. For the knowledge sharing module (KSM), we employ 2 layers of fully connected network with dimensions 768 in the TK-Layer. {We employ 3 transfer capsules. We experimented with 2 to 15 capsules and selected 3 based on the validation data accuracy.} % \zixuan{(we do hyper-parameter search range from 2 to 15, among which 3 gives the best result in validation set for both ASC and DSC.)} {\color{red}(?? what happens if more or less?)}. %{\color{red}(?? need update this subsection as we do not use dynamic routing anymore. Need to consider both DSC and ASC.)}
% The dynamic routing is repeated for 3 iterations.
For the task specific module (TSM), {we use 
%employ the embedding with \zixuan{final and hidden layer is not "embeddings", may use "size of"} 
2000 dimensions as the final and the hidden layers of the TSM.} The task ID embeddings have 2000 dimensions. A fully connected layer with softmax output is used as the classification heads in the last layer of the BERT, together with the categorical cross-entropy loss.
% We use 400 for $s_{\max}$ in Eq.~\ref{eq:smax} \bing{HAT details have been removed},
% \hu{remind what this is?} 
dropout of 0.5 between fully connected layers. The training of BERT, Adapter-BERT and CTR follows that of \cite{DBLP:conf/naacl/XuLSY19}. We adopt $\text{BERT}_{\textbf{BASE}}$ (uncased). The maximum input length is set to 128 which is long enough for both ASC and DSC. %of the sum of sentence and aspect is set to 128 for ASC \zixuan{and the maximum length for DSC is set to 128}\bing{?? same}. 
We use Adam optimizer and set the learning rate to 3e-5.
% Since the two SemEval datasets have much more examples than the other datasets, 
{We use 10 epochs for SemEval datasets and 30 epochs for the other datasets in the ASC application. For DSC, we use 20 epochs. For 20News, we use 10 epochs. These are selected based on the validation results. %The number of epochs are chosen based on results from the validation data. 
The batch size is set to 32 for all cases.}
% \zixuan{they do not use in BERT experiment but CL methods}. %\bing{?? why?}. 
For all the other CL baselines, we use the code provided by their authors and adapt them for text classification. We also adopt their original parameters, including learning rate, early stopping, and batch size. % \bing{Add something about the 20News data}

% Number of parameters and training time are given in \textit{Supplementary}. 

\begin{table}[t]
\centering
\resizebox{\columnwidth}{!}{
\begin{tabular}{ccc||cccccccccc}
\specialrule{.2em}{.1em}{.1em}
\multirow{2}{*}{Scenarios} & \multirow{2}{*}{Category} & \multirow{2}{*}{Model} & \multicolumn{2}{c}{ASC} & \multicolumn{2}{c}{DSC (small)} & \multicolumn{2}{c}{DSC (full)} & \multicolumn{2}{c}{20News} & \multicolumn{2}{c}{20News (FR)}\\
 &  &  & Acc. & MF1 & Acc. & MF1 & Acc. & MF1 & Acc. & MF1 & Acc. & MF1\\ \specialrule{.1em}{.05em}{.05em}
\multirow{4}{*}{\begin{tabular}[c]{@{}c@{}} Non-continual \\ Learning (SDL)\end{tabular}} & \multicolumn{1}{l}
{BERT} & MTL &	91.91 &	88.11 &	85.05 &	84.03 &	89.77 &	89.28 &	96.77 &	96.77 & --- & ---\\
& \multicolumn{1}{l}{BERT} & SDL & 85.84 & 76.35 & 78.04 & 74.17 & 87.84 & 86.80 & 96.49 & 96.48 & --- & ---\\
  & BERT (Frozen) & SDL & 78.14 & 58.13 & 73.88 & 67.97 &  85.34 & 80.17 & 96.49 & 96.48 & --- & ---\\
 & \multicolumn{1}{l}{Adapter-BERT} & SDL & 85.96 & 78.07 & 76.31 & 71.04 & 88.30 & 87.31 & 96.20 & 96.19 & --- & ---\\
 & \multicolumn{1}{l}{W2V} & SDL & 77.01 & 51.89 & 62.06	& 53.80 & 69.57 & 65.51 & 94.72	& 94.72 & --- & ---\\

 \cline{1-13}
\multirow{25}{*}{\begin{tabular}[c]{@{}c@{}}Continual \\      Learning (CL)\end{tabular}} & \multicolumn{1}{l}{BERT} & NFH & 49.60 & 43.08 & 73.87 & 69.44 & 73.08 & 71.81 & 52.50 & 39.22 & 24.29	& 30.52\\
 & BERT (Frozen) & NFH & 85.51 & 76.64 & 83.12 & 79.23 &  61.88 & 45.79 & 83.28 & 81.81 & 8.76 & 9.73\\ 
 & \multicolumn{1}{l}{Adapter-BERT} & NFH & 54.03 & 44.81 & 63.76 &  53.95 & 64.94 & 63.40 & 68.29 & 61.70 & 30.59 & 3.79\\
 & \multicolumn{1}{l}{W2V} & NFH & 82.69 & 73.56 & 65.16	& 57.48 & 70.40 & 68.03 & 90.74 & 90.59 & 4.30 & 4.47\\  
 \cline{2-13}
 & \multirow{6}{*}{BERT   (frozen)}
 & L2 & 56.04 & 38.40 & 59.17 & 48.39 & 69.80 & 62.63 & 72.14 & 65.39 & 24.57 & 32.05\\
 &  & A-GEM &  86.06 & 78.44 & 59.33 & 45.94 & 70.67 &  61.77 & 93.31 & 92.95 & 4.09 & 4.48\\
  &  & DER++ & 84.27 & 75.08 & 72.29 & 66.28 & 86.70 & 85.46 & 60.44 &  49.67 & 10.54 & 12.16 \\
 &  & KAN & 85.49 & 77.38 & 77.27 & 72.34 & 82.32 & 81.23 & 73.07 & 69.97 & 15.52 & 18.87\\
 &  & SRK & 84.76 & 78.52 & 78.58 & 76.03 & 83.99 & 82.66 & 79.64 & 77.89 & 12.06 & 13.97\\
 &  & EWC & 86.37 & 74.52 & 82.38 & 78.41 & 72.77 & 65.76 & 80.26 & 78.60 & 3.50 & 3.03 \\
 &  & UCL & 83.89 & 74.82 & 80.12 & 74.13 & 74.76 & 69.48 & 94.65 & 94.63 & 0.48 & 0.48\\
 &  & OWM & 87.02 & 79.31 & 58.07 & 42.63 & 86.30 & 85.36 & 84.54 & 82.73 & 13.80 & 15.81\\
 &  & HAT & 86.74 & 78.16 & 79.48 & 72.78 & 87.29 & 86.14 & 93.51 & 92.93 & 2.26 & 2.89\\
  &  & CAT & 83.68	& 68.64 & 67.41 & 56.22  & 87.34 & 86.51  & 95.17 & 95.16 & 0.80 & 0.81\\
 \cline{2-13}
 & \multirow{4}{*}{Adapter-BERT} & L2 & 63.97 & 52.43 & 67.26 & 62.76 & 73.03 & 71.50 & 69.56 & 65.50 & 23.12	& 27.39\\
 &  & A-GEM & 45.88 & 28.21 & 62.89 & 55.96 & 71.22 & 69.94 & 60.29 & 50.40 & 40.22 & 51.20\\
  &  & DER++ & 47.63 & 35.54 & 70.52 & 63.56 & 59.67 & 57.82 & 58.95 & 49.58 & 36.39 & 45.30\\
 &  & EWC & 56.30 & 49.58 & 58.23 & 51.03 & 62.69 & 61.51 & 61.86 & 53.94 & 37.79 & 46.58\\
 &  & UCL & 64.46 & 36.64 & 48.30 & 32.07 & 57.06 & 55.86 & 51.75 & 36.06 & 4.70 & 6.60\\
 &  & OWM & 72.99 & 66.51 & 73.97 & 71.96 & 85.46 & 84.57 & 71.10 & 66.25 & 27.38 & 32.76\\
 &  & HAT & 86.14 & 78.52 & 80.83 & 78.41 & 88.00 & 87.26 & 95.22 & 95.21 & 0.33 & 0.34\\
%  &  & CAT &  &  &  &  &  &  & &  \\
 \cline{2-13}

 & \multirow{6}{*}{W2V} & L2 & 60.36 & 39.13 & 54.34 & 43.19 & 57.71 & 48.00 & 59.54 & 54.40 & 7.83 & 11.89\\
 &  & A-GEM &  81.33 & 63.35 & 69.80 & 60.07 & 77.67 & 70.75 & 90.72 & 90.60 & 3.94	& 4.08\\
 &  & DER++ & 83.27 & 69.93 & 77.51 & 73.13 &  74.79 & 66.68 & 89.28 & 89.19 & 4.32 & 4.42\\
 &  & KAN & 72.06 & 40.01 & 57.13 & 43.75 & 69.35 & 64.78 & 57.92 & 51.65 & 20.98 & 27.02\\
 &  & SRK & 71.01 & 39.63 & 64.47 & 55.93 & 69.65 & 65.25 & 61.07 & 58.47 & 7.26  & 8.81\\
 &  & EWC & 84.16 & 72.29 & 64.82 & 57.20 & 70.00 & 65.11 & 91.86 & 91.80 & 2.64  & 2.71\\
 &  & UCL & 84.41 & 75.99 & 56.23 & 41.34 & 70.56 & 67.01 & 90.61 & 90.46 & 4.53  & 4.70\\
 &  & OWM & 82.70 & 71.18 & 53.40 & 38.44 & 67.15 & 65.42 & 71.97 & 68.75 & 24.00 & 27.56\\
 &  & HAT & 80.83 & 63.63 & 62.57 & 50.83 & 69.75 & 65.44 & 67.73 & 64.43 & 26.04 & 29.70\\
 &  & CAT & 76.28 & 54.65 & 55.19 & 35.28 & 79.58 & 75.99 & 70.38 & 68.04 & 24.37 & 26.95\\
 \cline{2-13}
& \multicolumn{2}{c||}{B-CL}
&  88.29 & 81.40 & 82.01 & 80.63 & 79.76 & 76.51 & 95.07 & 95.04 & 0.58 & 0.59\\
& \multicolumn{2}{c||}{LAMOL}  & 88.91 &	80.59 &	89.12 &	86.58 &	92.11 &	91.72 &	66.13 &	45.74 & 20.03 & 16.60\\
 \cline{2-13}
 & \multicolumn{2}{c||}{CTR (forward)} & 87.89 & 80.25 & 83.75 & 82.55 & \text{89.86} & \text{89.16} & \text{95.63} & \text{95.62} & --- & ---\\
 & \multicolumn{2}{c||}{\text{CTR}} & {\text{89.47}} & {\text{83.62}} & {\text{84.34}} & {\text{83.29}} & {89.31} & {88.75} & 95.25 & 95.23 & 0.42 & 0.43\\
\specialrule{.1em}{.05em}{.05em}
% \vspace{-4mm}
\end{tabular}
}
\caption{Accuracy (Acc.) and Macro-F1 (MF1), averaged over 5 random sequences.  {\color{black}The architectures of SRK, KAN and CAT cannot work with Adapter-BERT.} ``---'' means not applicable. {\color{black}Due to the limited space, {\textit{standard deviation}, \textit{\textit{running time}} and \textit{\textit{network size}}} are given in Supplementary.} %\bing{blod?}\zixuan{seems you remove the bold in results. Or you mean bold what are in supplementary? <-I think we should them}} % \zixuan{note backward is not necessary better}
% {\color{red}The detailed table of statistical analysis are given in \textit{Supplementary}.}
}
% \hu{rename Adapter to Adapter-MLP ? }
% \hu{summarize the effects of knowledge transfer and forgetting? do we need statistical testing?}
% \vspace{-4mm}
%The number in bold in each row is the best result of the row.}
% \hu{
% it seems how to use the embeddings are not important but how to do CL is more important; (re-group into NL/NFH with different sub-rows? 
%TODO: add citation to each baseline if possible.)
%}
\vspace{-6mm}
\label{tab:overall_results}
\end{table}

\vspace{-1mm}
\subsection{Evaluation Results and Analysis}
\label{sec:results}
Since the order of the tasks in a sequence may impact the final results, we randomly sampled 5 task sequences and averaged their results.
We compute both accuracy and Macro-F1, %over 3 classes of opinion polarities for ASC and 2 classes for DSC, %(2 classes for datasets other than SemEval), 
where Macro-F1 is the primary metric as the imbalanced classes (in ASC) introduce biases on accuracy. {\color{black}Table~\ref{tab:overall_results} gives the average results of all the 19 tasks (or datasets) for ASC and 10 tasks (or datasets) for DSC and 20News over the 5 random task sequences.} 

% {for ASC and DSC, accordingly.}} % all systems on 5 random task sequences of 19 tasks (or datasets). % The detailed results for individual tasks are given in the supplementary. \zixuan{Individual Results have been removed from supplementary}
% Column 1 gives the scenarios of whether it is for continual learning. Column 2 gives the category the systems belong to. Column 3 gives the specific name of the system. 

\textbf{Overall Performance.} 
% \zixuan{inconsistent name convention}
%The \textit{overall} accuracy results of all tasks for our datasets in 
Table~\ref{tab:overall_results} shows that CTR clearly outperforms all baselines. % We discuss the details below:
% \hu{Do we want to have a subsubsection for "effectiveness of forgetting avoidance"? we have one for "knowledge transfer"}

(1).~For the standalone (SDL) baselines, BERT {(with fine-tuning)} and Adapter-BERT perform similarly. % with Adapter-BERT slightly better. %\zixuan{only the adapter blocks are trainable in Adapter-BERT, can we sTask-CLl call it fine-tunning?}\bing{Fine-tuning has been used. Without fine-tuning, the results are significantly poorer.} 
W2V {and BERT (Frozen) are} poorer, which is understandable.

{\color{black}(2). Comparing SDL (standalone learning) and NFH (continual learning with no forgetting handling), we see NFH is much better than SDL for W2V {and BERT (Frozen)}. This indicates ASC {and DSC} tasks have similarities and thus shared knowledge. Catastrophic forgetting (CF) is not a major issue for W2V {and BERT (Frozen)}. However, for 20News, NFH variants have serious CF. % compared to the SDL variants. 
NFH with BERT (fine-tuning) is much worse than SDL and Adapter-BERT, which we explained in Introduction. % this is because BERT with fine-tuning learns highly task specific knowledge \cite{DBLP:journals/corr/abs-2004-14448}. While this is desirable for SDL, it is bad for NFH because task specific knowledge is hard to share (and should not be transferred) across tasks. Then NFH % but easily results in negative-transfer and thus 
% causes serious forgetting (CF).
} 

(3). Unlike BERT and Adapter-BERT, CTR does very well in both forgetting avoidance and knowledge transfer (outperforming all baselines). %\bing{does CTR also use contextual (Frozen BERT), what happened to Adapter-BERT here? What are the results of baselines with word2vec?}. 
For baselines, EWC, UCL, OWM and HAT, although they perform better than NFH, they are significantly poorer than CTR as they don't have methods to encourage knowledge transfer for ASC and DSC. KAN and SRK {do knowledge transfer but they are weaker than many other CL methods.} They perform very poorly for 20News as they have limited ability to overcome CF. CAT works well with large datasets, but is weak for small datasets. % \bing{more discussions after CAT experiments}.

(4). CTR's improvements over SDL variants for DSC (large) is less than for DCS (small). This is understandable because when the training data is large, learning a separate model is already good, and knowledge transfer is less important. 

(5). Compared with the SDL results, we can see that CTR has the least forgetting on 20News. 

{\color{black}(6). Compared to B-CL, CTR is markedly better in knowledge transfer. The forgetting rates (FR) of B-CL and CTR are both low. {\color{black}The comparison is in fact the same as comparing dynamics routing and transfer routing. We can see that the proposed transfer routing is dramatically better than dynamic routing for knowledge transfer. %, which already mentioned in Sec. 3: "with a new \textit{transfer routing} algorithm to explicitly identify transferable features/knowledge from previous tasks to transfer to the new task.
Additionally, transfer routing eliminates the need for hyper-parameter tuning on the number of iterations of dynamic routing~\cite{sabour2017dynamic} to update the agreements.} 
% does not need multiple iterations 

(7). CTR outperforms LAMOL in ASC and 20News even with the less powerful BERT model that CTR adopts. LAMOL outperforms BERT-based MTL in DSC. This may be because LAMOL is based on GPT-2, which is known to be more powerful than BERT (also shown in the LAMOL paper). For 20News, since its tasks are very different/dissimilar, there is little shared knowledge. Dealing with CF is the main issue. LAMOL has serious forgetting as its FR values show. % , which may generate poor pseudo examples and result in forgetting. CTR has almost no forgetting.

(8). The results of MTL (the upper bound) are only slightly better than CTR, which again shows that CTR is highly effective in overcoming forgetting and encouraging knowledge transfer. }
% do knowledge transfer but they are for document-level sentiment classification. They are weak, even weaker than other CL methods. 

% (4) Comparing BERT and Adapter-BERT, we can see that Adapter-BERT outperforms BERT in both NL and NFH, indicating the advantage of employing Adapter-BERT.  

% \zixuan{different size of training data has been removed}

\textbf{Effectiveness of Knowledge Transfer.}
{\color{black}Forward transfer is defined as the forward performance (\textbf{CTR(forward)} in Table~\ref{tab:overall_results}) subtracting the standalone (SDL) result. CTR(forward) is the test accuracy or MF1 of each task when it was first learned. Backward transfer is defined as the difference between the backward performance (\textbf{CTR} in Table~\ref{tab:overall_results}, the final result after all tasks are learned) and the forward performance.  % Comparing the results of SDL with the results of forward transfer, we can see whether forward transfer is effective. By comparing the forward transfer result with the backward transfer result, we can see whether the backward transfer can improve further. 
{The average results of CTR (forward) and {CTR} are given in Table \ref{tab:overall_results}. We observe that forward transfer of CTR is highly effective for the three datasets with similar tasks. For DSC, the less the data, the more effective is the transfer, which is reasonable.} 
% (forward results for other CL baselines are given in \textit{Supplementary}). 
{Backward transfer improves the accuracy and MF1 of ASC and DSC (small). For DSC (full), it is slightly weaker and for 20News, it is also slightly weaker due to a very small amount of forgetting}, but the less than 0.0055 CF is negligible. {\color{black} Note that in~\cite{Lopez2017gradient,riemer2018learning}, forward transfer is measured by comparing the test results of the new task on the current learned network and a random initialized network before/without training the new task (like zero-shot). This method indicates whether the learned network contains some useful knowledge for the new task. However, it does not tell how much forward transfer actually happens after learning the new task, which is more important and is what our method measures. }}
% \bing{we may compute the forgetting rate} % \zixuan{following is not true now. we may want a different conclusion or choose not to give conclusion yet}and we see CTR's forward result outperforms all baselines' forward results ). For backward transfer, CTR also improves the performance.
% After the two transfers, accuracy gains have been made by CTR. 

{\textbf{Overcoming Forgetting.} To validate CTR's effectiveness in dealing forgetting with a sequence of dissimilar tasks, {\color{black}we compute the \textit{Forgetting Rate} \cite{DBLP:conf/cvpr/LiuSLSS20},
$\text{FR} = \frac{1}{t-1}\sum_{i=1}^{t-1}A_{i,i} - A_{t,i}$,
where $A_{i,i}$ is the forward accuracy of task $i$ and $A_{t,i}$ is the accuracy of task $i$ after training the last task $t$. We average over all tasks except the last one because the last task obviously has no forgetting. {\color{black}We report the forgetting rate FR (averaged over 5 random task  sequences) for the 20News data on the two evaluation metrics in the last two columns of Table~\ref{tab:overall_results} (the other two datasets are mainly for knowledge transfer). CTR has very low FR values which indicate very little forgetting.} % just like HAT with CTR having the highest accuracy and macro-F1 scores.
} % \zixuan{May be also mention the smaller FR, the better}

\textbf{Ablation Study.} 
{The results of ablation experiments are in Table \ref{tab:ablation_results}. ``-KSM;-TSM'' means that the knowledge sharing module (KSM) and the task specific module (TSM) are not used, and {we simply deploy the Adapter-BERT.} ``-KSM'' means that the knowledge sharing module (KSM) is not used. ``-TSM'' means that the task specific module (TSM) is not used. {``-TR/KSM'' means that the task router (TR) in KSM is not used. We directly send the transferable representation $v_{j|i}^{\text{tran}(t)}$ to the next layer. ``dynamic routing'' means that we replace our transfer routing (-(SE\&TR)/KSM) with dynamic routing~\cite{sabour2017dynamic}, which is one of the most popular routing algorithms in capsule networks.}  % i.e., replacing our transfer routing with dynamic routing. 
% \zixuan{``-(TR)/KSM; + Dynamic'' means we replace the TR module with dynamic routing (i.e. the $v_{j|i}^{\text{tran}(t)}$ is sent to dynamic routing).}
%\zixuan{the "FFN" and "attention" may be confused. we may need a better name} 
{\color{black}``on top'' means adding a CL-plugin on top of BERT. ``after the first FF layer'' means adding only one CL-plugin there in BERT. ''after the other two FF layers'' means adding only one CL-plugin module there in BERT.
Table \ref{tab:ablation_results} shows that the full CTR system gives the best results, indicating every component contributes to the model and other options of adding CL-plugins are all poorer.}}}

\begin{table}[]
\centering
\resizebox{0.9\columnwidth}{!}{
\begin{tabular}{c||cccccccc}
\specialrule{.2em}{.1em}{.1em}
\multirow{2}{*}{Model} & \multicolumn{2}{c}{ASC} & \multicolumn{2}{c}{DSC (small)}  & \multicolumn{2}{c}{DSC (full)} & \multicolumn{2}{c}{20News} \\
 & Acc. & MF1 & Acc. & MF1 & Acc. & MF1  & Acc. & MF1\\
\specialrule{.1em}{.05em}{.05em}
CTR (-KSM;-TSM)  &  0.5403 & 0.4481 & 0.6376 & 0.5395 & 0.6494 & 0.6340 & 0.6829 & 0.6170 \\
CTR (-TSM)  &  0.8312 & 0.7107 & 0.7085 & 0.6759 & 0.8545	& 0.8380 & 0.8275 & 0.8064\\
CTR (-KSM)  &  0.8614 & 0.7852 & 0.8083 & 0.7841 & 0.8800 & 0.8726 & 0.9522 & 0.9521\\
CTR (-TR/KSM)  &  0.8819 & 0.8155 & 0.8244 & 0.8119 & 0.8831 & 0.8762 & 0.9476 & 0.9469 \\
% CTR (-TR/KSM; +Dynamic)  &  0.8845 & 0.8257 & 0.8225 & 0.7987 \\
CTR (dynamic routing) (B-CL)
&  0.8829 & 0.8140 & 0.8201 & 0.8063 & 0.7976 & 0.7651 & 0.9507 & 0.9504 \\
CTR (on top) 
&  0.8135 & 0.6390 & 0.7301 & 0.6875 & 0.8266 & 0.8105 & 0.8927	 & 0.8920 \\
CTR (after the first FF layer) 
&  0.8741 & 0.8014 & 0.8300 & 0.8183 & 0.8699 & 0.8596 & 0.9381 & 0.9373 \\
CTR (after the other two FF layers) 
&  0.8662 & 0.7863 & 0.8269 & 0.8161 & 0.8714 & 0.8612  & 0.9339 & 0.9316 \\
\textbf{CTR} &  \textbf{0.8947}
 & \textbf{0.8362}
 &  \textbf{0.8434} & \textbf{0.8329} & \textbf{0.8931} & \textbf{0.8875} & \textbf{0.9525} & \textbf{0.9523}\\
\specialrule{.1em}{.05em}{.05em}
\end{tabular}
}
\caption{Ablation experiment results. %\zixuan{added new ablation} %\bing{include 20News results}\zixuan{included}
}
\label{tab:ablation_results}
\vspace{-4mm}
\end{table}

%\subsection{Error Analysis}
% \hu{how some patterns of sequence of tasks, break down your 5 random and see why some are worse because of some ordering? visualization of masks with domain labels, and capsule hidden space? } \zixuan{Do we need it? }

\section{Conclusion}
{\color{black}This paper studied task continual learning (Task-CL) using the pre-trained model BERT to achieve both CF presentation and knowledge transfer.} 
It proposed a novel technique called CTR to leverage the pre-trained BERT for CL. The key component of CTR is the CL-plugin inserted in 
BERT. A CL-plugin is a capsule network with a new transfer routing mechanism to encourage knowledge transfer among tasks and also to isolate task-specific knowledge to avoid forgetting. Experimental results using three NLP applications showed that CTR markedly improves the performance of both the new task and the old tasks via knowledge transfer and is also effective at overcoming catastrophic forgetting. {\color{black}One limitation of our work is the efficiency due to the use of capsules. Capsules try to represent a group of neurons in a vector reflecting properties of an entity. In NLP, an entity is a sentence/document which contains many tokens (e.g., 128) and features (e.g. 768 in $\text{BERT}_{\text{BASE}}$). Grouping them makes the capsule very large (e.g., 128 $\times$ 768), which slows down training.}

\section*{Acknowledgments}
{\color{black}This work was supported in part by two National Science Foundation (NSF) grants (IIS-1910424 and IIS-1838770), a DARPA Contract HR001120C0023, and a Northrop Grumman research gift.} 

\bibliography{neurips_2021}
\bibliographystyle{abbrv}

\end{document}